\documentclass[runningheads]{llncs}

\usepackage{microtype}
\usepackage{graphicx}
\usepackage{booktabs} % for professional tables
\usepackage{hyperref}
\usepackage[ruled,vlined]{algorithm2e}
\usepackage{caption}
\usepackage{nicefrac}  % compact symbols for 1/2, etc.
\usepackage{multirow}
\usepackage{array}
\usepackage{mathtools}
\usepackage{amsfonts} % blackboard math symbols
\usepackage{amsmath}
\usepackage{bbm}
\usepackage{amssymb}
\usepackage{dsfont}
\usepackage{float}
\usepackage{xurl}
\usepackage{subcaption}
\usepackage{pgfplots}
\usetikzlibrary{plotmarks}
\usepgfplotslibrary{fillbetween}
\pgfplotsset{compat=1.15}
\pgfplotsset{
    discard if not/.style 2 args={
        x filter/.code={
            \edef\tempa{\thisrow{#1}}
            \edef\tempb{#2}
            \ifx\tempa\tempb
            \else
                
            \fi
        }
    }
}

\newcommand{\hide}[1]{}
\newcommand{\repeatthanks}{\textsuperscript{\thefootnote}}

\usepackage{xspace}
\usepackage{color}

\newcommand*{\ie}{\emph{i.e.},\@\xspace}
\newcommand*{\etc}{\emph{etc.}\@\xspace}

\newcommand*{\modelname}{\texttt{CheXRelNet}\@\xspace}

\begin{document}

\title{\modelname: An Anatomy-Aware Model for Tracking Longitudinal Relationships between Chest X-Rays}
\titlerunning{Anatomy Aware Model for Longitudinal Relationships Change in CXRs}

\author{
Gaurang Karwande \thanks{Equal Contribution, with authors listed in alphabetical order.}\inst{1}
\and
Amarachi B. Mbakwe \repeatthanks\inst{1}
\and
Joy T. Wu\inst{2,3}
\and 
Leo A. Celi\inst{4,5,6}
\and
Mehdi Moradi\inst{7} 
\and
Ismini Lourentzou\inst{1}}

%index{Karwande, Gaurang}
%index{Mbakwe, Amarachi B}
%index{Wu, Joy T}
%index{Celi, Leo A}
%index{Moradi, Mehdi}
%index{Lourentzou, Ismini}

\authorrunning{G. Karwande et. al.}
\institute{Virginia Tech \email{\{gaurangajitk,bmamarachi,ilourentzou\}@vt.edu} \and
Standford Medicine \email{joytywu@stanford.edu}  \and
IBM Research \and
Institute for Medical Engineering and Science, Massachusetts Institute of Technology \and
Beth Israel Deaconess Medical Center \and
Harvard T.H. Chan School of Public Health \email{lceli@mit.edu}\and
University of British Columbia \email{moradi@ece.ubc.ca}
}

\maketitle            

\begin{abstract}
Despite the progress in utilizing deep learning to automate chest radiograph interpretation and disease diagnosis tasks, change between sequential Chest X-rays (CXRs) has received limited attention. Monitoring the progression of pathologies that are visualized through chest imaging poses several challenges in anatomical motion estimation and image registration, \ie spatially aligning the two images and modeling temporal dynamics in change detection.
In this work, we propose \modelname, a neural model that can track longitudinal pathology change relations between two CXRs. \modelname incorporates local and global visual features, utilizes inter-image and intra-image anatomical information, and learns dependencies between anatomical region attributes, to accurately predict disease change for a pair of CXRs. Experimental results on the Chest ImaGenome dataset show increased downstream performance compared to baselines. Code is available at \url{https://github.com/PLAN-Lab/ChexRelNet}.

\end{abstract}

\section{Introduction}
Medical imaging research has experienced tremendous growth over the past years, spurred by continuous AI advancements~\cite{Tang2020,majkowska2020chest,kiradoo2018role}, and in particular in the development of specialized digital devices and neural medical imaging architectures. Chest radiography is one of the most performed diagnostic examinations worldwide. The demand for chest radiography has increased the radiologists' workload. As manually interpreting Chest X-rays (CXRs) and radiology reports can be time-consuming, these challenges contribute to the delays in detecting findings and providing exemplary patient clinical management plans. Though there has been substantial progress in radiology such as disease diagnostics \cite{articlediagnostics,10.3389/frai.2020.583427,articlediag,unknown}, medical image segmentation \cite{MAITY2022103398,articleseg,s21020369,Reamaroon2020}, \etc, more complex reasoning tasks remain fairly unexplored.
For example, despite significant progress in the application of machine learning in chest radiograph medical diagnosis, detecting longitudinal change between CXRs has attracted limited attention. 
Yet, understanding whether the patient's condition has deteriorated or improved is crucial to guide the physician's decision-making and determine the patient's clinical management.

Automating this process is a challenging task. At times, differences between x-rays might go undetected, hindering early detection of disease progression that requires an immediate change in treatment plans. Previous work tackles change between longitudinal patient visits and evaluates the severity of diseases at each time point on a continuous scope on osteoarthritis in knee radiographs and retinopathy of prematurity in retinal photographs \cite{Li2020}. 
Other works target longitudinal disease tracking and outcome prediction severity for COVID-19 pulmonary diseases \cite{PMID:33928256}, by calculating a severity score for pulmonary x-rays via computing the Euclidean distance between each of the normal images and the image of interest. 
In addition, geometric correlation maps have been used to study the CXR longitudinal change detection problem~\cite{inbook}, in which feature maps are extracted from CXR pairs and their matching scores are used to generate a geometric correlation map that can detect map-specific patterns showing lesion change. 
However, these works rely on global image information. To the best of our knowledge, no prior work considers capturing correlations among anatomical regions and findings when modeling change between CXRs. Yet, localizing pathologies to anatomy is critical for the radiologists' reasoning and reporting process, where correlations between image findings and anatomical regions can help narrow down potential diagnoses. 

The development of imaging models that track progress or retrogression between CXRs findings or diseases remains still an open issue. 
Therefore, in this work, we propose  \modelname, an anatomy-aware neural model that utilizes information from anatomical regions, learns their intra-image and inter-image dependencies with a graph attention network, and combines the localized region features with global image-level features to accurately capture anatomical location semantics for each finding when performing longitudinal relation comparison between CXR exams for a variety of anatomical findings.

The contributions of this work are summarized as follows: 1) we introduce \modelname, an anatomy-aware model for tracking longitudinal relations between CXRs. The proposed model utilizes both local and global anatomical information to output accurate localized comparisons between two sequential CXR examinations, 2) we propose a graph construction to capture correlations between anatomical regions from a pair of CXRs, and 3) we conduct experimental analysis to demonstrate that our proposed \modelname model outperforms baselines. Finally, 4) we perform transfer learning experiments to test the generalization capabilities of our model across pathologies.
\section{Methodology}\label{sec:method}
Let $\mathcal{C} = \{({x}_i,{x}'_i)\}_{i=i}^{N}$ be the set of CXR image pairs. Each image $x_i$ has $k$ anatomical regions. In addition, each image is associated with a set of labels $\mathcal{Y}_i = \{y_{i,m}\}^{M}_{m=1}$, $y_{i,m} \in \{0,1\}$ indicating whether the label for pathology $m$ appears in image $x_i$ or not, and each pair $({x}_i,{x}'_i)$ is associated with a set of labels $\mathcal{Z} = \{z_{i,m}\}_{m=1}^M, z_{i,m} \in \{0,1\}$ indicating whether the pathology $m$ appearing in the image pair has improved or worsened. 
The goal is to design a model that compares the two images and predicts their labels as accurately as possible for an unseen image pair $(x,x')$ and a wide range of pathologies. This is achieved by utilizing (i) the correlation among anatomical region features from the images $x_i, x'_i$, \ie $R = f(x)$ and $R'=f(x')$, $R,R' \in \mathbb{R}^{k \times d}$, where $k$ is the number of anatomical regions, each embedded into a row vector with dimensionality $d$ (extracted by a pretrained feature extractor $f$) and (ii) the correlation among anatomical regions between the two images in the pair.

Given the initial training set of anatomical region representations $\{(R_i,R'_i)\}_{i=1}^N$, we define a normalized adjacency matrix $A \in \mathbb{R} ^{2k \times 2k}$ that captures intra-image and inter-image region correlations. 
\begin{figure}[t!]
    \includegraphics[width=1.0\columnwidth]{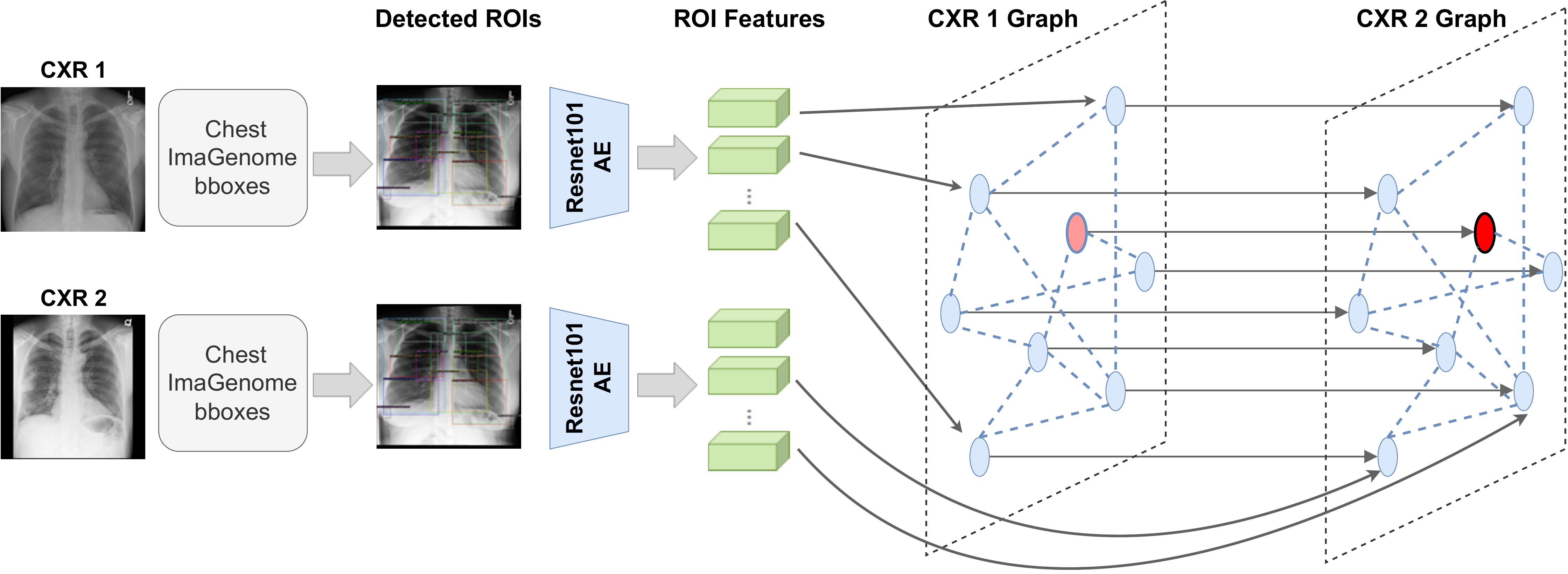}
    \caption{Graph Construction overview. Detected anatomical regions of interest (ROIs) are fed into a ResNet101 pretrained autoencoder to extract their corresponding visual features, formulating initial node representations $\mathcal{V}_i,\mathcal{V}'_i$ for CXR graphs $\mathcal{G}_i=(\mathcal{V}_i, \mathcal{E})$ and $\mathcal{G}'_i=(\mathcal{V}'_i, \mathcal{E})$. Here, $\mathcal{E}$ is constructed based on intra-image region-disease co-occurrence. Moreoever, the nodes of the two graphs are connected via a set $\mathcal{\tilde{E}}$ of directed edges indicating inter-image relations.}
    \label{fig:Graph}
\end{figure}

The intra-image correlations corresponding to the two $k \times k$ diagonal blocks of $A$ are constructed based on the region-disease co-occurrence ~\cite{agu2021anaxnet}, \ie the number of times two anatomical regions co-occur with the same disease or finding in the set of images $R_i, R'_i, i=1,\ldots,N$. Each of these $k \times k$ co-occurrence blocks can be computed via the Jaccard similarity
\begin{equation}
\label{eq:Jaccard similarity}
J(r_s, r_t) = \frac{1}{M}\sum_{m=1}^{M}\frac{|\mathcal{Y}_{s,m} \cap  \mathcal{Y}_{t,m}|}{|\mathcal{Y}_{s,m}
\cup  \mathcal{Y}_{t,m}|}.
\end{equation}
Here, $r_s$ represents an anatomical region, $\mathcal{Y}_s^m$ is the set of disease labels for region $r_s$ and pathology $m$ across all images and $\cap, \cup$ denote the intersection and union over multi-sets. 
To overcome the shortcomings of the label co-occurrence construction tendency to overfit the training data, a filtering threshold $\tau$ is adopted, \ie
\begin{equation}
\label{eq: filtering}
A_{st} =
    \begin{cases}
      1 & \text{if $J(R_s, R_t)$ $\geq$ $\tau$}\\
      0 & \text{if $J(R_s, R_t)$ $<$ $\tau$}
    \end{cases}.
\end{equation}
Here, $A_{st}$ corresponds to an element of one of the two diagonal blocks. We further note that the two diagonal blocks are identical. 

The inter-image correlations correspond to the two off-diagonal $k \times k$ blocks of the adjacency matrix $A$, and they are chosen to indicate a relationship between the same anatomical regions of every pair of images. More precisely, we set $A_{st} = \mathbbm{1}\{t=s+k\}$ for $s=1,\ldots,k$.
The rationale of this adjacency matrix definition is that $A$ will be associated with every pair $(x_i, x'_i)$ and will capture useful inter-image correlations and local intra-image region-level correlations.
More precisely, the upper $k \times k$ diagonal block is associated with image $x_i$, forming a graph $G_i = (\mathcal{V}_i, \mathcal{E})$ with nodes being the vector representations of the $k$ anatomical regions of image $x_i$. Similarly, the lower $k \times k$ diagonal block is associated with image $x'_i$, forming a graph $G'_i = (\mathcal{V}'_i, \mathcal{E})$ as before. Finally, the upper $k \times k$ off-diagonal block indicates a set of edges $\mathcal{\tilde{E}}$ between the same regions of images $x_i, x'_i$. This graph construction is also depicted in Figure \ref{fig:Graph}.

To capture global and local dependencies between anatomical regions, we utilize a graph attention network (GAT)~\cite{velivckovic2018graph}
$Z_i = g(R_i,A) \in \mathbb{R}^{k \times d}$ to update $R_i$ as follows:

\begin{equation}
R_i^{(t+1)} = \alpha^{(t)}_{i,i}W_1^{(t)}R_i^{(t)} +
\sum_{j \in \mathcal{N}(i)} \alpha^{(t)}_{i,j}W^{(t)}_1R_j^{(t)},
\end{equation}
where $W_1 \in \mathbb{R} ^{d \times d}$ is a learned weight matrix,  $\mathcal{N}(i)$ denotes the neighborhood of $x_i$, $t$ is the number of stacked GAT layers, and $\alpha_{i,j}$ are the attention coefficients computed as 
\begin{equation}
 \alpha^{(t)}_{i,j} =
\frac{
\exp\left(\mathrm{LeakyReLU}\left(\mathbf{a}^{\top}
\left[W_1^{(t)}{R}_i^{(t)}; W_1^{(t)}{R}_j^{(t)}\right]
\right)\right)}
{\sum\limits_{k \in \mathcal{N}(i) \cup \{ i \}}
\exp\left(\mathrm{LeakyReLU}\left(\mathbf{a}^{\top}
\left[W_1^{(t)}{R}_i^{(t)} ; W_1^{(t)}{R}_j^{(t)}\right]
\right)\right)}
\end{equation}

Here, $\mathbf{a}$ is a learned weight vector, and $;$ denotes concatenation. The final region representations are computed by a weighted combination of the neighbor vector representations, scaled by their attention scores
\begin{equation}
R_i^{(t+1)}=\phi\left(\sum_{j\in \mathcal{N}(i)} {\alpha^{(t)}_{ij} R_j^{(t)} }\right),
\end{equation}
where $\phi(\cdot)$ is a non-linear transformation. Given the past history of the patient, a medical expert has enough information to direct the majority of their focus on a particular region within a CXR. In Figure \ref{fig:Graph}, the node highlighted in red corresponds to the physician-designated focus region $k^* \in [1,k]$ for the particular CXR examination. 
We extract the node embedding corresponding to the focus region of $x'_i$ for each CXR image pair and forward this embedding to the final dense classification layer. 
Specifically, for a focus-region $k^* \in [1,k]$, the extracted node embedding $R'_i \in \mathbb{R}^{d}$ is given by,
\begin{equation}
R'_i = {R'_i}^{(t+1)} \mathbbm{1}\{k=k^*\}.
\end{equation}

\begin{figure}[t!]
    \centering
    \includegraphics[width=0.7\columnwidth]{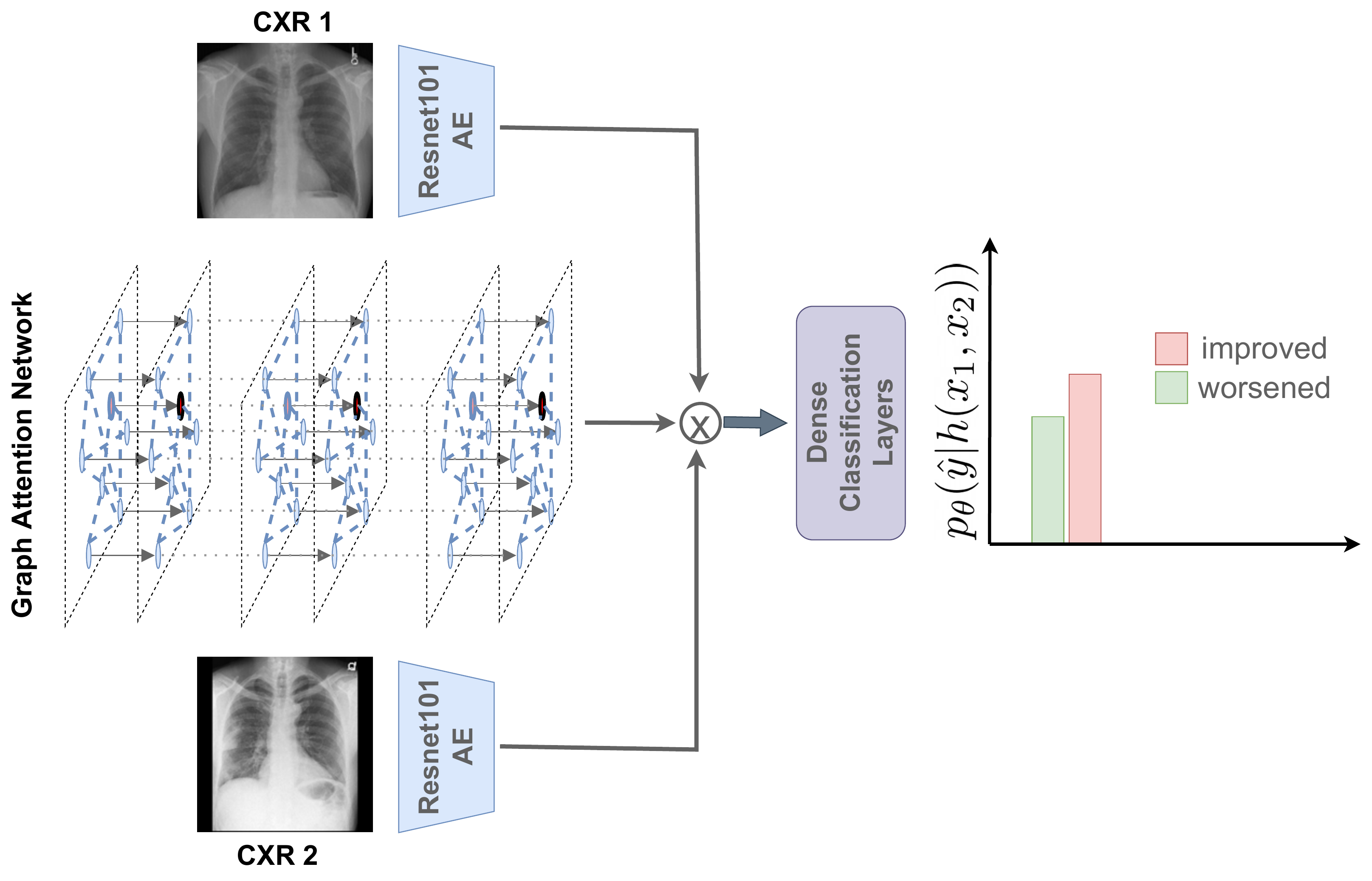}
    \caption{Classification module. The constructed graph, \ie the anatomical regions of interest (ROIs) vector representations and the corresponding adjacency matrix, is passed through a graph attention network that learns ROI inter-dependencies, essentially capturing local ROI information. Global image-level representations extracted from a pretrained ResNet101 autoencoder model are concatenated with the ROI learned representations and are passed through a final dense classification layer. The model is trained end-to-end with a cross-entropy classification loss.}
    \label{fig:Classificationmodule}
\end{figure}

To capture global image-level information, each image in pair $(x_i, x'_i)$ is encoded into two $d$-dimensional vectors by utilizing the pretrained feature extractor $f$, \ie $ Q_i = f(x_i)$ and $Q'_i = f(x'_i)$, $Q_i,Q'_i  \in \mathbb{R}^{d}$. The final prediction is computed via  
\begin{equation}
\hat{y} = \left[ R'_i ; Q_i; Q'_i \right] W_2^T,
\end{equation}
where $;$ denotes the concatenation of the local region-level and global image-level features, $W_2 \in \mathbb{R}^{3d \times M} $ is a fully connected layer that obtains the label predictions. The network is trained with a multi-label cross-entropy classification loss
\begin{equation}
L = \frac{1}{N} \sum_{i=1}^{N} \sum_{m=1}^{M} z_{i,m} log(\sigma(\hat{z}_{i,m})) + (1-z_{i,m})log(1-\sigma(\hat{z}_{i,m})),
\label{losseqn}
\end{equation}
where $\sigma$ is the sigmoid function and $\{\hat{y}_i^m, y_i^m\} \in \mathbb{R} ^{M}$ are the model prediction and the ground truth for example $x_i$, respectively. Figure \ref{fig:Classificationmodule} presents an overview of the model architecture.
\section{Experiments}\label{experiments}
\begin{table}[t!]
\caption{Dataset Characteristics. \# Image Pairs (number of comparison CXR pairs) and \# Bboxes (number of bounding boxes) and \# Training Pairs (number of training comparison CXR pairs) per pathology label. Each pathology is indexed with a pathology ID (first column).}
 \centering
 \resizebox{\textwidth}{!}{
\begin{tabular}{ccccc}
 \toprule
 \textbf{{Pathology ID}} & \textbf{{Description}}  & \textbf{{\# Image Pairs}} & \textbf{{\# Bboxes}} & \textbf{{\# Training Pairs}} \\
 \midrule
D1 & Lung Opacity & 32,524 & 455,336 & 22,620 \\
D2 & Pleural Effusion & 13,122 & 183,708 & 9,192\\
D3 & Atelectasis & 9,660 & 135,240 & 6,922\\
D4 & Enlarged Cardiac Silhouette & 1,958 & 3,916 & 1,384\\
D5 & Pulmonary Edema/Hazy Opacity & 12,090 & 169,260 & 8,424 \\
D6 & Pneumothorax & 2,728 & 38,192 & 1,930 \\
D7 & Consolidation & 3,332 & 46,648 & 2,310 \\
D8 & Fluid Overload/Heart Failure & 674 & 9,436 & 132 \\
D9 & Pneumonia & 3,814 & 53,396 & 2,590 \\
 \midrule 
All 9 Pathologies & \textbf{{Total}} & 79,902  & 1,095,132 & 55,504 \\ 
\bottomrule
 \end{tabular}
}
\label{tab:dataset_description}
\end{table}

\subsection{Dataset}
The proposed \modelname model is trained and evaluated on the \textsc{{Chest ImaGenome}} dataset \cite{wu2021chest}. This dataset was generated by locally labeling $242,072$ frontal MIMIC-CXRs~\cite{johnson2019mimic} (AP or PA view) automatically through a combination of rule-based text analysis and atlas-based bounding box extraction techniques~\cite{wu2020automatic,wu2020ai}. \textsc{{Chest ImaGenome}} represents the connections of each CXR annotation as an anatomy-centered scene graph, following a radiologist-constructed CXR ontology. The dataset contains $1,256$ combinations of relation annotations between $29$ CXR anatomical locations and their attributes structured as one scene-graph per image, and about $670,000$ localized comparison relations between the anatomical locations across sequential exams. In this work, we utilize the localized comparison relations data that involves cross-image relations for the $9$ pathologies. Each comparison relation in the \textsc{{Chest ImaGenome}} dataset consists of the DICOM identifiers of the two CXRs being compared, the particular pathological finding observed in those two CXRs, the anatomical region of interest on which the radiologist's comparison is focused, and the corresponding comparison label. In addition to comparison relations, the \textsc{{Chest ImaGenome}} dataset also provides bounding box information for extracting individual anatomical regions from the CXRs, viz. `Left Lung', `Cardiac Silhouette', \etc For each of the $242,072$ frontal MIMIC-CXRs a list of anatomical regions (bboxes) is provided, as well as the corresponding euclidean coordinates for each bounding box. We utilize these coordinates to crop different anatomical regions within a CXR.
There are a total of $122,444$ unique comparisons in the dataset, of which $79,902$ have at least one of the nine selected pathology labels, in addition to regions detected by the object detection pipeline and the overall comparison relation. For each image, except for those with the pathology label as `Enlarged Cardiac Silhouette', $7$ of the most frequently occurring anatomical regions were extracted. For the pathology label `Enlarged Cardiac Silhouette', the dataset provides only one corresponding bounding box. Table \ref{tab:dataset_description} shows high-level data statistics.

\subsection{Baselines}
We compare the \modelname model against the following baselines: 1) \textbf{Local} model: we utilize a previously proposed siamese network trained on cropped ROIs, encoded with a pre-trained ResNet101 autoencoder and passed through a dense layer and a final classification layer~\cite{wu2021chest}. This model essentially only looks at the corresponding anatomical regions and considers neither global information nor intra-region dependencies.
2) \textbf{Global} model: we also design a siamese architecture that encodes the entire CXR as opposed to only the cropped ROIs in the Local model. Apart from the input being a full image rather than an ROI, the model architecture is the same as the Local model. Hence, this baseline incorporates the global information but does not take into consideration the anatomical region of interest nor explicitly models inter-region dependencies.
These two siamese models serve as baseline methods to contrast the effectiveness of \modelname, which not only is location-aware but can also explicitly model both inter-region and intra-image CXR dependencies.

\begin{table}[t!]
\centering
\caption{Comparison against baselines (accuracy).}
\resizebox{\textwidth}{!}{%
\begin{tabular}{p{2cm}*{10}{p{0.8cm}<{\centering}}}
\toprule
Method & D1 &  D2 & D3 & D4 & D5 & D6 & D7 & D8 & D9 & \textbf{All}  \\
\midrule
Local & 0.59 & 0.53 & 0.60 & 0.47 & 0.56 & 0.46 & 0.61 & 0.47 & 0.63 & 0.60\\
Global & 0.67 & \textbf{0.69} & 0.64 & 0.74 & 0.71 & 0.50 & 0.65 & 0.69 & 0.67 & 0.67\\
\modelname & 0.67 & 0.68 & \textbf{0.66} &\textbf{0.75} & 0.71 & \textbf{0.52} & \textbf{0.67} & \textbf{0.73} & 0.67 & \textbf{0.68}\\
\bottomrule 
\end{tabular}
}
\label{tab:results}
\end{table}

\subsection{Implementation Details}
To train each model, we use the train/validation/testing splits and detected ROIs provided by \textsc{Chest ImaGenome}. 
For each image within the comparison pair, we crop the image ROIs and resize them to $224 \times 224$. Each cropped ROI is then embedded into a vector with $2048$ dimensions, by utilizing a pre-trained ResNet101 autoencoder \cite{Cohen2022xrv} that is trained on several medical imaging datasets, e.g., NIH, CheXpert, and MIMIC datasets, \etc \cite{irvin2019chexpert,johnson2019mimic,wang2017chestx}. The same autoencoder is utilized for the baseline models.
The co-occurrence matrix threshold is set to $0.5$.
Our model is a 2-layer graph neural network with $2048$ and $1024$ neurons per layer, in the first and second layers respectively. There are $5$ and $3$ multi-head-attentions in each respective layer. The output from the graph attention network is concatenated with the global information and then passed through two dense layers of sizes $768$ and $128$ respectively. We train the network using Adam \cite{kingma2015adam} optimizer for $200$ epochs, with a $0.8e^{-3}$ initial learning rate set \cite{wilson2001need} and a batch size of $32$. To avoid overfitting, we utilize early stopping with $11$ patience and gradient clipping that is set to $0.1$. In addition, we use $0.5$ Dropout~\cite{JMLR:v15:srivastava14a} and a learning rate decay factor of $0.3$ with the patience threshold set to $4$. The model is implemented by utilizing the PyTorch \cite{NEURIPS2019_9015} and pytorch\_geometric \cite{Fey/Lenssen/2019} deep learning frameworks. The evaluation metric is accuracy and results are reported over six experimental trials.

\subsection{Experimental Results}
Results are summarized in Table \ref{tab:results}. \modelname achieves a mean accuracy of $0.683$ (SD=$0.0024$), while the Local model has  $0.602$ mean accuracy (SD=$0.0059$) and the Global model has $0.672$ mean accuracy (SD=$0.0046$) over six trials.
We observe that the Local model is generally underperforming, and it is most likely limited because it focuses on a specific anatomical region and completely neglects the global information. In contrast, radiologists often take into consideration more than one anatomical region when drawing inferences from CXRs.
The Global model is a lot more effective than the Local one, and incorporating global information boosts the prediction accuracy. Yet, the Global model is also limited as it focuses on the entire image but fails to consider the relationships among anatomical regions. 
We additionally perform statistical significance tests, \ie an unpaired t-test $(p=0.049)$ and a one-tailed t-test $(p=0.018)$ comparing \modelname and the Global baseline. These t-test results verify that the \modelname and Global baseline predictions follow distinct distributions and that the improvement in accuracy is significant at $p < 0.05$. Additional experiments w.r.t. model architectures and model capacity (number of parameters) are reported in the supplementary. Overall, \modelname improves upon the Global model's prediction accuracy by modeling the inter-image and intra-image region correlations and attending to the anatomical regions of interest.

We also perform a transfer learning experiment wherein we train \modelname on a set of diseases and test performance on a different set of diseases~\cite{chao2016empirical,xian2017zero}. Specifically, we train \modelname on a subset of the data with `Pneumothorax', `Consolidation', `Fluid Overload/Heart Failure', `Pneumonia' (D6-D9) pathologies, and test on the following pathology labels that are unseen during training: `Lung Opacity', `Pleural Effusion', `Atelectasis', `Enlarged Cardiac Silhouette', and `Pulmonary Edema/Hazy Opacity' (D1-D5). Results are reported in Table \ref{tab:results_zeroShot}. We perform this experiment on individual unseen pathology labels as well as on sets of multiple unseen pathology labels. We observe that our model can generalize well to unseen pathology labels. We can attribute this to the incorporation of both local and global information during training. The model is learning associations between different anatomical regions and therefore can identify complex bio-markers associated with the progression of pathologies.
\begin{table}[t!]
\centering
\caption{Transfer learning evaluation against baselines (accuracy). Models are trained on D6-D9 and tested on unseen pathologies (D1-D5). SetA consists of unseen pathologies  \{D1, D2\}. SetB consists of unseen pathology labels, \{D3, D4\}. Set C consists of all unseen pathology labels \{D1,D2,D3,D4,D5\}.}
\resizebox{0.9\textwidth}{!}{%
\begin{tabular}{p{3cm}*{10}{p{0.8cm}<{\centering}}}
%\begin{tabular}{ccccccccc}
\toprule
Method & D1 &  D2 & D3 & D4 & D5 & SetA & SetB & SetC  \\
\midrule
Local & 0.56 & 0.49 & 0.54 & 0.49 & 0.55 & 0.54 & 0.55 & 0.54\\
Global & 0.61 & \textbf{0.63} & 0.60 & 0.65 & 0.63 & 0.61 & 0.63 & 0.62\\
\modelname (ours) & \textbf{0.64} & 0.60 & \textbf{0.61} & \textbf{0.68} & \textbf{0.67} & \textbf{0.63} & \textbf{0.64} & \textbf{0.64}\\
\bottomrule 
\end{tabular}
}
\label{tab:results_zeroShot}
\end{table}

\section{Conclusion}\label{sec:concl}
CXRs are commonly repeatedly requested in the clinical workflow to assess for a myriad of attributes. Diagnosis and monitoring are typically performed through comparisons of sequential CXR images, both in in-patient and outpatient settings. Given a patient with two sequential CXR exams, the goal of this work is to automatically evaluate disease change. To this end, we describe a methodology for localized relation comparisons between CXR images. The proposed \modelname  fuses global image-level information, local intra-image region-level correlations, and inter-image correlations. Experimental results show that \modelname outperforms baselines in both traditional and transfer learning settings. 
As a result, our method provides necessary components for monitoring the progression of pathologies that are visualized through chest imaging. 
In the future, we hope to expand our work to model disease progression among several sequential CXRs, incorporate additional temporal context information and physiological data ~\cite{wu2021chest,karargyris2021creation} and account for the time interval variability found in longitudinal imaging records. Finally, future work can adapt the proposed methodology to other medical imaging tasks, and include interpretability mechanisms.

\bibliographystyle{splncs04}
\bibliography{biblio}
\newpage
\section{Supplementary Material}\label{sec:concl}
\subsection{Qualitative Results}
\begin{figure}[h!]
    \centering
    \includegraphics[width=0.8\columnwidth]{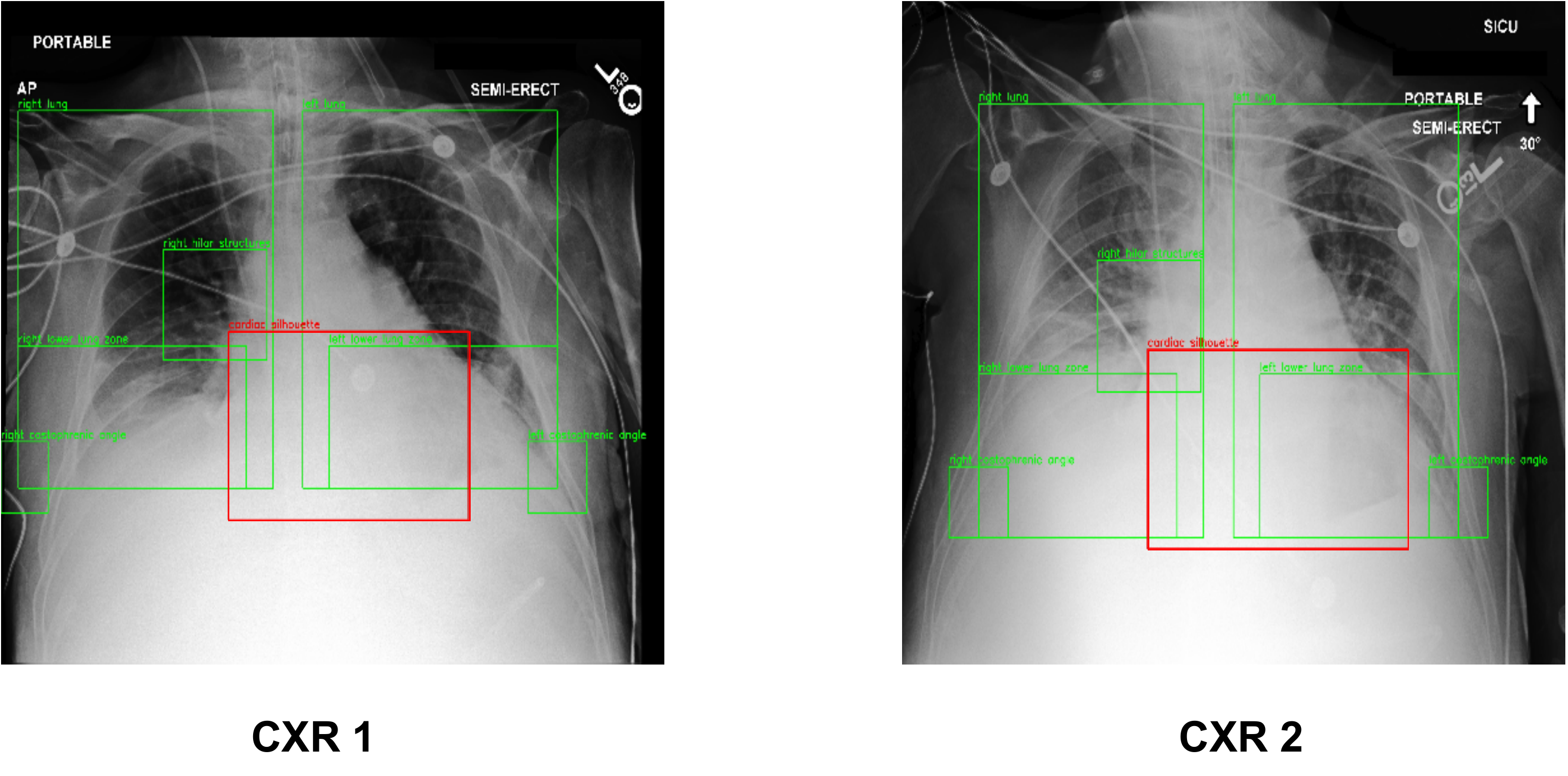}
    \caption{Qualitative results for pathology D8, class: Worsened}
    %graph construction and the graph structure is based on the adjacency matrix co-occurence computation joined with the directed edges from the previous CXR to the current CXR.}
    \label{fig:qualitative_1}
\end{figure}

\begin{figure}[h!]
    \centering
    \includegraphics[width=0.8\columnwidth]{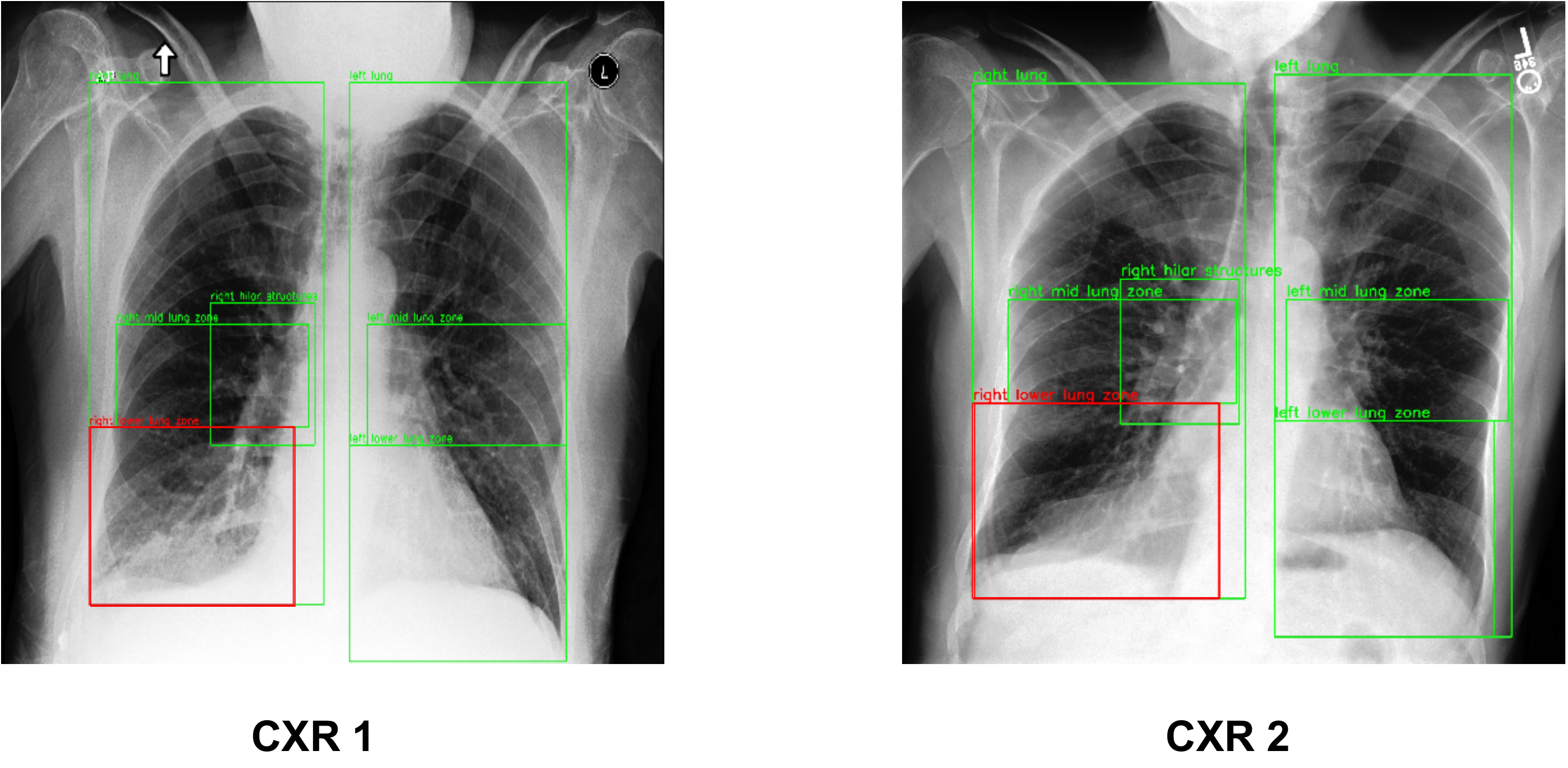}
    \caption{Qualitative results for pathology D9, class: Improved}
    %graph construction and the graph structure is based on the adjacency matrix co-occurence computation joined with the directed edges from the previous CXR to the current CXR.}
    \label{fig:qualitative_2}
\end{figure}

We visualize the model predictions for different pathologies. Figure \ref{fig:qualitative_1} showcases an input image pair for the pathology label `Fluid Overload/ Heart Failure' where there has been a worsening in the patient's condition. For this particular pair, the anatomical region of interest (ROI) is `Cardiac Silhouette', which is depicted with a red bounding box. Other anatomical regions that our model takes into consideration when making predictions are shown in green bounding boxes. The previous and current CXRs are named `CXR 1' and `CXR 2', respectively. Upon close inspection, we can see there are subtle changes within the ROI as well as in other parts of the CXR. There is increased haziness in the Left and Right Lungs, and minute changes in the Cardiac Silhouette. Similarly, Figure~\ref{fig:qualitative_2}, depicts the input image pair for the pathology label `Pneumonia', and the case where there has been an improvement in the patient's condition. For this particular pair, the anatomical ROI is `Right Lower Lung Zone'. In addition to changes within the ROI, there are significant improvements in the regions `Left Mid Lung Zone' and `Left Lower Lung Zone'. The Local model focuses only on the ROI, whereas the Global model focuses only on the entire image. Hence, both of these fail in making correct predictions over these images. Our \modelname model builds associations between various regions and hence is able to factor in the minute changes across the entire anatomy while making predictions.

\subsection{Pathology-specific Models}
In this experiment, we train models that are specific to a given pathology label. Hence, unlike Table~\ref{tab:results}, where the model is trained jointly for all nine pathology labels and later tested over each individual pathology, here we train and test our model on the same pathology label. Results are shown in Table~\ref{tab:results_singleDisease}. From these results, we can infer that \modelname is comparable to or outperforms the Local and Global baselines for five out of nine pathologies. For the pathologies `Pleural Effusion' (D2), `Consolidation' (D7), and `Pneumonia' (D9), the difference in accuracy is greater than or equal to $5 \%$.
Compared to the results in Table~\ref{tab:results}, another interesting observation is that when our model is trained for all pathologies, the classification accuracy is higher. We attribute this observation to the presence of a consistent pattern of disease progression across all pathologies. Hence, the model trained on all pathologies is able to generalize better and is more consistent than the one trained on a single pathology.

\begin{table}[t!]
\centering
\caption{Pathology-specific comparison of \modelname against baselines.}
\resizebox{\textwidth}{!}{%
\begin{tabular}{p{3cm}*{10}{p{0.8cm}<{\centering}}}
%\begin{tabular}{cccccccccc}
\toprule
Method & D1 &  D2 & D3 & D4 & D5 & D6 & D7 & D8 & D9 & AVG \\
\midrule
Local & 0.63 & 0.55 & 0.59 & 0.62 & 0.68 & \textbf{0.53} & 0.60 & 0.45 & 0.63 & 0.59\\
Global & \textbf{0.68} & 0.64 & \textbf{0.61} & 0.69 & \textbf{0.70} & 0.49 & 0.59 & \textbf{0.69} & 0.58 & 0.63\\
\modelname (ours) & 0.67 & \textbf{0.69} & 0.61 & \textbf{0.71} & \textbf{0.70} & 0.49 & \textbf{0.67} & 0.65 & \textbf{0.65} & \textbf{0.65}\\
\bottomrule 
\end{tabular}
}
\label{tab:results_singleDisease}
\vspace{-0.3cm}
\end{table}

\subsection{Multi-class Classification with `No Change'}

\begin{table}[t!]
\centering
\caption{Comparison against baselines with `no change' samples.}
\resizebox{\textwidth}{!}{%
\begin{tabular}{p{3cm}*{10}{p{0.8cm}<{\centering}}}
%\begin{tabular}{ccccccccccc}
\toprule
Method & D1 &  D2 & D3 & D4 & D5 & D6 & D7 & D8 & D9 & \textbf{All}  \\
\midrule
Local & 0.41 & 0.37 & 0.41 & 0.29 & 0.37 & \textbf{0.37} & \textbf{0.49} & 0.29 & 0.42 & 0.43\\
Global & 0.45 & \textbf{0.47} & \textbf{0.44} & 0.48 & 0.48 & 0.36 & 0.47 & \textbf{0.50} & 0.43 & 0.45\\
\modelname (ours) & \textbf{0.49} & \textbf{0.47} & \textbf{0.44} & \textbf{0.49} & \textbf{0.49} & 0.36 & 0.47 & 0.44 & \textbf{0.47} & \textbf{0.47}\\
\bottomrule 
\end{tabular}
}
\label{tab:results_noChange}
\vspace{-0.3cm}
\end{table}
Table~\ref{tab:results_noChange} shows the results when we repeat the original experiment with samples wherein there is no change in the disease progression. Albeit the accuracy drops considerably for all models, \modelname outperforms baselines. Further investigations are required to make the model robust to `no change' samples.

\begin{table}[t!]
\centering
\caption{Ablation study on model structure and capacity.}
\resizebox{1.0\textwidth}{!}{
\begin{tabular}{@{}lcccccccccc@{}}
    \toprule
    \textbf{Model} &
      \multicolumn{3}{c}{\textbf{Local}} &
      \multicolumn{3}{c}{\textbf{Global}} &
      \multicolumn{3}{c}{\textbf{\modelname}} &\\
      \cmidrule(r){2-4}\cmidrule(l){5-7} \cmidrule(l){8-11}
      & {Type A} & {Type B} & {Type C} & {Type A} & {Type B} & {Type C} & {Type A} & {Type B} & {Type C} & {Type D} \\
      \midrule
    \#Parameters (M) & 25.6 & 34.7 & 41.4 & 25.6 & 34.7 & 41.4 & 38.6 & 27.8 & 54.3 & 28.9  \\
    Accuracy & 0.60 & 0.64 & 0.63 & 0.67 & 0.67 & 0.67 & 0.68 & 0.68 & 0.68 & 0.67  \\
    \bottomrule
  \end{tabular}
  }
\label{tab:ablation}
\end{table}
 % Chexrelnet deeper - 54.3 M parameters, 0.686113 accuracy

\subsection{Ablation Study: Model Architectures and Capacity}
We perform an ablation study to investigate if there exists a correlation between the performance and the model capacity (number of trainable parameters). Results are shown in Table~\ref{tab:ablation}. For all three models, the architectures named \textit{Type A} are the ones used throughout the study. The architectures named \textit{Type B} and \textit{Type C}, for the Local and Global baselines, have more dense layers and neurons. As for the graph models, \textit{Type B}, \textit{Type C} are the shallower and deeper versions of \modelname having 1 and 3 Graph Attention (GAT) layers, respectively, whereas \textit{Type A} is the actual \modelname that has 2 GAT layers. \textit{Type D} is the version of our \modelname model without attention and in that we replace the GAT layers with simpler Graph Convolution layers. The number of trainable parameters and corresponding accuracy is reported in Table~\ref{tab:ablation}. We can infer that the model performance is less influenced by the model capacity and that the graph neural network is the prominent differentiating factor.

\end{document}